\documentclass[letterpaper]{article}
\usepackage{iccc}

\usepackage{times}
\usepackage{helvet}
\usepackage{courier}

\usepackage{amsmath}
\usepackage{graphicx}
\usepackage{subfig}
\usepackage[hyphens]{url}
\usepackage{xcolor}

\title{Seeding Diversity into AI Art\\
Paper type: Technical Paper}
\author{Marvin Zammit, Antonios Liapis and Georgios N. Yannakakis\\
Institute of Digital Games, University of Malta, MSD2080, Malta\\
\{marvin.zammit,antonios.liapis,georgios.yannakakis\}@um.edu.mt
}
\begin{document} 
\maketitle
\begin{abstract}
\begin{quote}
This paper argues that generative art driven by conformance to a visual and/or semantic corpus lacks the necessary criteria to be considered creative. Among several issues identified in the literature, we focus on the fact that generative adversarial networks (GANs) that create a single image, in a vacuum, lack a concept of novelty regarding how their product differs from previously created ones. We envision that an algorithm that combines the novelty preservation mechanisms in evolutionary algorithms with the power of GANs can deliberately guide its creative process towards output that is both good and novel. In this paper, we use recent advances in image generation based on semantic prompts using OpenAI's CLIP model, interrupting the GAN's iterative process with short cycles of evolutionary divergent search. The results of evolution are then used to continue the GAN's iterative process; we hypothesise that this intervention will lead to more novel outputs. Testing our hypothesis using novelty search with local competition, a quality-diversity evolutionary algorithm that can increase visual diversity while maintaining quality in the form of adherence to the semantic prompt, we explore how different notions of visual diversity can affect both the process and the product of the algorithm. Results show that even a simplistic measure of visual diversity can help counter a drift towards similar images caused by the GAN. This first experiment opens a new direction for introducing higher intentionality and a more nuanced drive for GANs.
\end{quote}
\end{abstract}

\section{Introduction}\label{sec:introduction}

Visual art is among the most well-researched domains in computational creativity as it is perhaps the most recognisable among tasks which, when performed by humans, are deemed
creative \cite{ritchie2007empirical}.
Painting in any style or medium requires some degree of skill \cite{colton2008tripod}, and endowing machines with painting skill has a long and exciting history \cite{cohen2017aaron,painting_fool,deussen2015painting,machado2002nevar}. A watershed moment in this endeavour has been the advent of Generative Adversarial Networks (GANs) \cite{goodfellow2014generative}, which not only started to bridge the gap between human and machine performance but also allowed novices to generate compelling images without extensive technical knowledge, development effort, or access to specialised hardware. Generative art produced through deep learned models has taken the world by storm in the last five years. The strength of models trained in vast image databases in producing highly typical content, such as human faces, has led to an almost ubiquitous fascination by researchers, artists, laymen, media, and speculators. We follow \citeauthor{mccormack2019autonomy} (\citeyear{mccormack2019autonomy}) and refer to visuals generated via deep learning as ``AI Art'' in this paper.

As the general public became more interested in AI Art, a crucial component for the perception of creativity hinged on whether the software could explain in natural language the framing information regarding what it was trying to portray \cite{colton2011face}. While several GAN architectures addressed the generation of images from text prompts \cite{reed2016learning,zhang2017stackgan}, they performed well only in limited datasets and could not scale to generate visuals based on broader themes. The recent introduction of OpenAI's Dall-E \cite{ramesh2021zero} demonstrated an unprecedented high correspondence between a given text prompt and the generated image on different prompts. While neither the Dall-E model nor the training dataset have been publicly released at the time of writing, a pre-trained model of Contrastive Language-Image Pretraining (CLIP) is available \cite{radford2021learning}. The release of CLIP energised researchers and enthusiasts alike, leading to many open-source projects and twitter bots that take advantage of the links between semantics and images to produce more convincing AI Art, such as album titles and covers\footnote{\url{https://twitter.com/ai_metal_bot}}.

In the context of computational creativity, however, it would be easy to argue that images generated only to conform to the patterns of the corpus fall into ``mere generation'' \cite{ventura2016generation} and lack authenticity \cite{mccormack2019autonomy}. Using the criteria of \emph{novelty}, \emph{quality} and \emph{typicality} regarding products of a creative process \cite{ritchie2007empirical}, we argue that GANs and similar architectures target only typicality by conforming to patterns discovered in their training corpus. While we appreciate that there are several issues---such as intent and attribution \cite{mccormack2019autonomy}---that AI Art should address before it can be considered creative, we focus in this paper on the novelty of the product by endowing the algorithm with a way to assess and prioritise diversity in its generated output. 

While a product's novelty can be assessed in terms of past artefacts of the same type, we focus instead on contemporaneous novelty in a population of artefacts that are generated---iteratively---at the same time. While GANs are typically applied to generate a single image, our study also tests how diverse a population of images produced by GANs can be when the initial seeds are different. We take advantage of evolutionary algorithms that perform quality-diversity search \cite{pugh2016quality} and combine them with the power of deep learning through \emph{cycles of exploration and refinement}. Taking advantage of trained models of semantic-image similarities, we test this process of iterative refinement \cite{liapis2013world} when generating sets of images for five text prompts. This first experiment raises a number of questions regarding e.g. how image novelty can be assessed, and we test two different image metrics as both evolutionary goals and for analysing the quality of the final results.

\section{Background Technologies}\label{sec:background}

The proposed methodology combines complex, cutting-edge technologies of deep learning and divergent evolution. The relevant technologies and a high-level overview of their inner workings are presented below.

\subsection{OpenAI CLIP}\label{sec:background_clip}
OpenAI's CLIP is a supervised neural network architecture which associates images with corresponding text and vice versa, learning underlying concepts within each of these domains \cite{radford2021learning}. CLIP was released in January 2021 and quickly became popular for a wide variety of tasks, such as image classification \cite{CLIPclassification}, semantic image generation \cite{ramesh2021zero}, and captioning \cite{mokady2021clipcap}.

CLIP is essentially a zero-shot classifier, which was pre-trained using images and corresponding textual phrases scraped from the internet. The training dataset itself was not released but it contained $4\cdot10^8$ text-image pairs. The structure of CLIP consists of a Transformer-based model \cite{vaswani2017attention} which encodes the tokenised input text batches. For the image encoding, two different architectures were compared; a ResNET-D based model \cite{he2019bag} and a Vision Transformer (ViT) model \cite{ViT_dosovitskiy}. Batches of image-text pairs are encoded and cross-processed using contrastive learning \cite{van2018representation} in order to train the model to predict the probability that a given text input matches a given image or vice versa. 
The resulting trained models matched or outperformed some of the best classifiers when applied to broad datasets, but ranked worse on specific, narrow-domain datasets. The benefit of CLIP in the current work is that it can provide a singular cosine similarity score (we refer to this as CLIP score in this paper) between a textual prompt and an image, for any semantic prompt. This CLIP score has been used to assess generated images and predetermined text input, and thus to steer various methods of GAN image generation towards some predetermined text input \cite{styleGAN_NADA,kim2021diffusionclip}. These CLIP-guided image generation experiments are often performed by enthusiasts and are not published; however, many early contributions are available in online repositories\footnote{\url{https://github.com/lucidrains/big-sleep} and \url{https://colab.research.google.com/drive/1L8oL-vLJXVcRzCFbPwOoMkPKJ8-aYdPN}, among others.}. 

In practice, CLIP-guided image generation starts from a random fractal noise array as an image, and uses CLIP to generate its embedding. CLIP is also used to embed the input text prompt and the two sets of vectors are compared using cosine similarity, given by: 
\begin{equation}
similarity(\vec{t}_{image}, \vec{t}_{prompt}) = \frac{\vec{t}_{image} \cdot \vec{t}_{prompt}}{|\vec{t}_{image}|\cdot|\vec{t}_{prompt}|}
\label{eq:cos_sim}
\end{equation}
\noindent where $\vec{t}_{image}$ and $\vec{t}_{prompt}$ are the CLIP vector embeddings of the image and the text prompt respectively, and $|\vec{t}|$ denotes the magnitude of vector $\vec{t}$.

\subsection{Generative Adversarial Networks}\label{sec:background_gan}

Generative Adversarial Networks (GANs) were introduced by \citeauthor{goodfellow2014generative} (\citeyear{goodfellow2014generative}) and have since become an important milestone in AI Art. GANs consist of a generator which learns to generate artefacts within its domain, and a discriminator network which learns to distinguish between what the generator creates versus real artefacts. The output of the discriminator is used to train the generator, pitting their progress against each other, and resulting in greatly enhanced performance compared to previous techniques. Generally, the discriminator is discarded after training and only the generator is used for inference. This technique has been used extensively across different domains, for text to image generation \cite{brock2018large}, style transfer \cite{karras2019style}, super-resolution upscaling \cite{2017srgan}, and many more applications.

Vector Quantized Variational Autoencoders (VQVAE) are autoencoders which operate on image segments instead of individual pixels \cite{esser2021taming}. Their networks combine convolutional layers with transformer structures, capturing short-range feature interactions with the former and long-range ones with the latter. An image at the encoder input is converted into a sequence of segments which are stored in a discrete code book of representations. An image is thus compressed to a sequence of indices representing the position of each segment within the code book.

During VQVAE training, a GAN architecture is used (often referred to as a VQGAN) to learn the weights and biases of the encoding and decoding networks, and also to determine the code book entries which will be available for these processes. Therefore, the training data has a significant impact on the variety of images which can be encoded or decoded by a VQGAN. Specifically, images with features that bear the closest resemblance to the training data set will be compressed and decompressed more faithfully than images in a different domain. As discussed in the introduction, products of VQGANs therefore target \emph{typicality} \cite{ritchie2007empirical} with the training set above all else.

VQGANs enable an easy translation between an image and its latent vector representation, offering a way to manipulate images that can be combined with both CLIP evaluation and latent vector evolution. By applying backpropagation to the latent vector conditioned by the CLIP score of its corresponding image to a given text prompt, an image can be directed towards a suitable representation for the latter.

\subsection{Novelty Search with Local Competition}\label{sec:background_nslc}

Evolutionary computation has a long history and has proven to be powerful in numerical optimisation tasks \cite{dejong1997handbook}. However, it often struggles to discover appropriate individuals that can act as stepping stones for further improvements towards an objective. In deceptive fitness landscapes, such individuals may perform poorly in terms of the objective but may possess the necessary genotypic structure that can lead to highly fit individuals after a number of genetic operators are applied. Novelty as the (sole) evolutionary objective was introduced ``as a proxy for stepping stones'' \cite{lehman2008novelty}. Novelty search has shown great promise in many application domains such as robotics \cite{lehman2008novelty,lehman2011novelty}, game content generation \cite{liapis2015ecj,liapis2013world,liapis2013delenox} and generative art \cite{lehman2012impressiveness}. While most publications in this vein apply novelty search to neuroevolution \cite{stanley2002neat}, it can also be applied to other types of indirect \cite{liapis2016arcade,liapis2013delenox} or direct embryogenies \cite{liapis2015ecj}.

Emulating evolutionary processes in nature, applying local competition to the process of natural selection showed greater promise in some applications \cite{lehman2011novelty}. Local competition pits individuals against phenotypically similar individuals in the search space. This Novelty Search with Local Competition (NSLC) allowed diverse features to survive and serve as stepping stones towards better specimen, even if their performance was only optimal locally. It empowered individuals with diverse features to survive and evolve without being overpowered by better developed features in other individuals. 
In practice, NSLC operates as a multi-objective optimisation problem where one objective is increasing the \emph{novelty score} and the other objective is increasing the individual's \emph{local competition score}. Both scores compare an individual with its nearest neighbours in a behavioural space; these neighbours may be from the current population or from an archive of novel past individuals. The novelty archive is populated during evolution, with the most novel individuals in every generation being added to the archive. The novelty score is calculated via Eq.~\eqref{eq:novelty}, as the average distance of this individual with its nearest $k$ neighbours. The local competition score is calculated via Eq.~\eqref{eq:competition}, as the ratio of nearest $k$ neighbours that the individual outperforms in terms of fitness. Evidently, the algorithm hinges on two important parameters: the \emph{fitness metric} which affects the local competition score, and the \emph{distance metric} which affects the nearest neighbours being considered and the novelty score in general.
\begin{align}
n(i) &= \frac{1}{k}\sum_{j=1}^{k}d(i,\mu_j)\label{eq:novelty}\\
lc(i) &= \frac{1}{k}\sum_{j=1}^{k}o_f(i,\mu_j)\label{eq:competition}
\end{align}
\noindent where $d(x,y)$ is the behavioural distance between individuals $x$ and $y$ and depends on the domain under consideration, $\mu_j$ is the $j$-th nearest neighbour to $i$,  $o_f(x,y)$ is 1 if $f(x)>f(y)$ and 0 otherwise, where $f$ is the fitness function for the current problem. The current population and the novelty archive are used to find the nearest neighbours.

\section{Proposed Methodology}\label{sec:methodology}

At its core, our proposed methodology revolves around an alternating sequence of refining cycles (via backpropagation) and exploration cycles (via divergent evolution). Using CLIP-guided VQGANs in this instance, we describe the specific methods followed in each cycle below.

\subsection{Backpropagation Cycle}\label{sec:method_backprop}
In order to take advantage of the power of GAN architectures in converting seemingly random noise into visually appealing images, we use backpropagation-driven cycles to start the process and as a final step of refining an image before showing it to a human audience. 

The code used for semantic image generation is based on Pixray\footnote{\url{https://github.com/pixray/pixray}}, using pixel-based generation through VQGANs \cite{esser2021taming}. Details of the VQGAN technologies are in the Background section. For this paper we adopt a VQGAN pre-trained on the WikiArt dataset \cite{tan2016fineart}. WikiArt is a dataset of $81,444$ images of artistic creations (paintings, images) across many different art styles\footnote{\url{https://archive.org/details/wikiart-dataset}}. The images produced from the WikiArt-trained VQGAN are more illustrative and surreal rather than representational or photorealistic, which suits our goals of producing artefacts that observers would consider creative. Moreover, the images generated for each prompt were deemed to be more visually similar to each other than other models, when starting from different random seeds. 

The generated images have dimensions of $384$ by $384$ pixels, and the VQVAE model sectioned the images into blocks of $16$ by $16$ pixels, resulting in a latent vector of $576$ integers, each representing an index of the code book entry used to represent that block. Each integer's value range is $[0,16384]$, as part of the autoencoder's code book. 

\begin{figure}[t]
\centering
\includegraphics[width=0.48\textwidth]{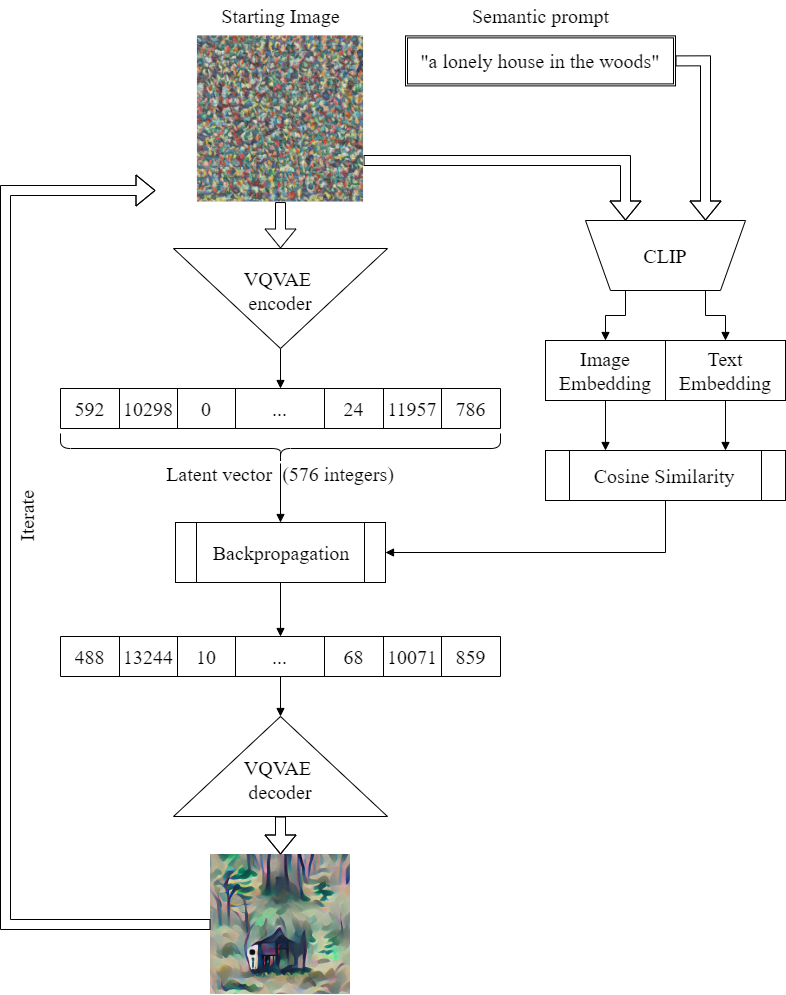}
\caption{GAN iterations guided by CLIP.}
\label{fig:gan_cycles}
\end{figure}

At the start of the experiment, we randomise each latent vector using a random fractal noise array, and use CLIP to generate its embedding. In subsequent iterations, we use the negated CLIP's cosine similarity of Eq.~\eqref{eq:cos_sim} as a loss function to guide the backpropagation process towards a latent vector which produces an image that better matches the semantic prompt. Since the latent vector consists of integers and is not compatible with the continuous requirement for gradient descent, the internal tensor representation of the vector within VQGAN (consisting of floating point numbers) is used to backpropagate the CLIP loss. Fig.~\ref{fig:gan_cycles} visualises this process.

\subsection{Exploration Cycle}\label{sec:method_exploration}

Exploration cycles are carried out via novelty search with local competition (NSLC) operating on the latent vector representing the image. The genotype (latent vector) consists of $576$ integers, ranging between $0$ and $16384$. Since each gene is an integer that is mapped in a very indirect way to some image segment, the evolutionary algorithm uses only mutation operators. In each mutation, 5\% of the individual's genes (chosen randomly) are replaced with random integers between $[0, 16384]$. This mutation rate was chosen based on initial trials, as it can create perceptible perturbations in the image without making it unrecognisable within one application of mutation (as is the case with higher mutation rates).

For this experiment, $k=15$ nearest individuals are considered for calculating both the novelty score and the local competition (LC) score, as per Eq.~\eqref{eq:novelty}-\eqref{eq:competition}. In each generation, the $e=3$ most novel individuals are added to the novelty archive. Note that the novelty archive starts empty at the start of each exploration cycle; there is no carryover from previous exploration cycles. The archive growth as the algorithm progresses increases the computational requirements, so this strategy of always adding a few individuals offered a good compromise between benefit and performance.

A Non-dominated Sorting Genetic Algorithm (NSGA-II) \cite{NSGA2} was used to process the resulting two metrics (novelty and LC score) as a multi-objective optimisation problem, using the Pymoo Python library \cite{pymoo}. A minimal Pareto front is calculated for the two objectives; individuals closest to this front are dominant over the remaining population and are selected for the next generation. If more individuals are required after exhausting those on the Pareto front, a next best Pareto front is calculated and a next set of individuals is selected from therein. If there are more individuals on the Pareto front than those required to survive, then individuals are selected to create sparsity in the objective space. The sparsity is based on the Manhattan distance between individuals within this space.

One of the major challenges in this work was defining diversity in the generated images for the purposes of NSLC. As noted in the Background section, the behavioural distance affects which neighbours are considered for both novelty and LC, and in turn affects how we envision novelty in the final product \cite{ritchie2007empirical}. It is a relatively easy task for a human to identify visual similarity between two images, but there are several challenges in quantifying similarity or diversity into a simple metric. In this work we compare image novelty by using two different approaches:

\subsubsection{Chromatic Diversity (HSV):}
With this approach, we hypothesise that the distribution of colours in the pixels reflect the diversity of the images \cite{machado2015complexity}. We consider the hue, saturation and brightness of each pixel and for any two generated images $I_1$ and $I_2$ we derive a diversity metric from their means and standard deviations as follows:
\begin{align}
m_1 &=\Delta\overline{b}=|\overline{b}_1-\overline{b}_2| \label{eq:imgeval_avgb} \\
m_2 &=\Delta{\sigma(b)}=|\sigma(b_1)-\sigma(b_2)| \label{eq:imgeval_stdb} \\
m_3 &=\Delta\overline{s}=|\overline{s}_1-\overline{s}_2| \label{eq:imgeval_avgs} \\
m_4 &=\Delta{\sigma(s)}=|\sigma(s_1)-\sigma(s_2)| \label{eq:imgeval_stds} \\
m_5 &=\Delta{\overline{h}}=|min[\overline{h}_1-\overline{h}_2,\overline{h}_1-(1-\overline{h}_2)]| \label{eq:imgeval_avgh} \\
m_6 &=\Delta{\sigma(h)}=|\sigma(h_1)-\sigma(h_2)| \label{eq:imgeval_stdh}
\end{align}
\noindent where $h$, $s$ and $b$ denote the hue, saturation and brightness, the means ($\overline{h},\overline{s},\overline{b}$) are taken across all the pixels in $I_1$ and $I_2$, and $\sigma$ denotes the standard deviation of these values. Note that since the hue value is cyclic, its mean and standard deviation were calculated as follows:
\begin{align}
    \overline{h}&=tan^{-1}\Bigg(\frac{\sum_{i=1}^N sin(h)}{\sum_{i=1}^N cos(h)}\Bigg)\\
    \sigma(h) &= \sqrt{\frac{\sum_{i=1}^N (min(h_i - \overline{h}, h_i - (1-\overline{h})))^2}{N-1}}
\end{align}  
\noindent where $N$ is the number of pixels in the image.

All $h, s, b$ values are normalised in the $[0,1]$ value range before the above calculations. We calculate the distance metric $d_{HSV}$ as the mean square value of the individual metrics $m_1{\ldots}m_6$.

\subsubsection{Visual Transformer Diversity (ViT):} 
Another way to assess diversity is based on the embeddings of pre-trained models. Transformers \cite{vaswani2017attention} have shown an outstanding performance when applied to image classification  \cite{ViT_dosovitskiy,ViT_Wu}. Within its layers, the model encodes information about different images in its training set, and uses it to discern different images. We utilise this encoded information with a ViT model pre-trained on the ImageNet data set \cite{deng2009imagenet}, and stripping its last layer. Since the last layer of ViT is used for image classification, by removing it we retain a latent vector of $768$ floating point values  for each processed image. We calculate a value of diversity ($d_{ViT}$) by taking the Euclidean distance between the latent vectors of two images.

\section{Experiment}\label{sec:results}

In order to assess how our envisioned algorithm that combines latent variable evolution \cite{bontrager2018deepmaster} towards novelty with GANs, the following section reports our findings when producing novel sets of images for different semantic prompts. We first describe our choice of prompts and parameter setup, followed by a quantitative analysis of both the process and the final product, and conclude with a qualitative view of the resulting images. 

\subsection{Protocol}\label{sec:results_protocol}

\begin{figure}[t]
\centering
\includegraphics[width=\columnwidth]{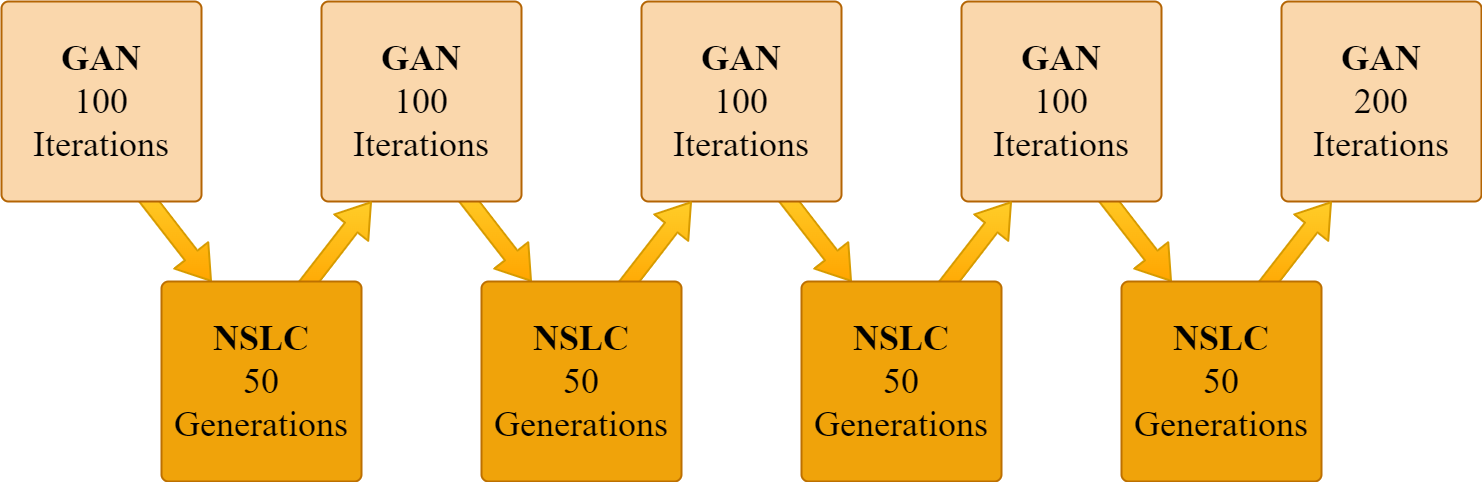}
\caption{Structure of the experiments alternating between GAN and evolutionary NSLC cycles.}
\label{fig:exp_cycles}
\end{figure}

For the purposes of demonstrating our proposed methodology in a visual creativity task, we use the Pixray image generation system which leverages pretrained VQVAE models. Importantly, we wish to explore how the method operates in a variety of settings while still being able to compare with existing research. To facilitate this, we test five semantic prompts (SP) used by the community\footnote{\url{https://github.com/lucidrains/big-sleep}}:
\begin{itemize}
\item a lonely house in the woods (SP1)
\item a pyramid made of ice (SP2)
\item artificial intelligence (SP3)
\item cosmic love and attention (SP4)
\item fire in the sky (SP5)
\end{itemize}

For this experiment, we generate a population of 50 images by running Pixray for a total of 600 iterations. The initial population consists of latent vectors encoded from a set of randomly generated fractal noise images. The same initial population of images is used in all tested variations of our algorithms, across all prompts. To establish our GAN baseline (GAN-BSL), we run the process uninterrupted for each initial latent vector for 600 iterations in order to collect the final population. Initial experiments showed that at 600 iterations the composition of the image is stable, and although more iterations will refine it, the image does not change much. For our NSLC experiments, we interrupt the GAN process after 100, 200, 300 and 400 iterations and take the latent vectors of the images at that point to produce an initial population for NSLC; NSLC evolves for 50 generations, guided by either ViT (NSLC-ViT experiment) or HSV (NSLC-HSV experiment) distance metrics, and the final evolved population is then used to continue the GAN process (until interrupted again). The process is clarified in Figure \ref{fig:exp_cycles}.

Evaluating the novelty or quality of the generated output is not straightforward \cite{ritchie2007empirical}. For the purposes of this paper, we align these notions with the quality-diversity characterisations of NSLC and use the following performance metrics to compare the different algorithms:
\begin{itemize}
\item \textbf{mean fitness} based on the CLIP score  across all 50 images in the population.
\item \textbf{mean ViT diversity} calculated as the average ViT distance from the nearest 15 neighbours per individual, averaged across all 50 images in the population. Note that for this metric only the current population is considered for finding nearest neighbours (no archive).
\item \textbf{mean HSV diversity} is calculated identically to mean ViT novelty using the HSV metric for measuring distance and finding nearest neighbours.
\end{itemize}

\subsection{Numerical Results}\label{sec:results_numerical}

\begin{figure*}[!t]
\centering
\subfloat[Mean population fitness]
{\includegraphics[width=0.32\textwidth]{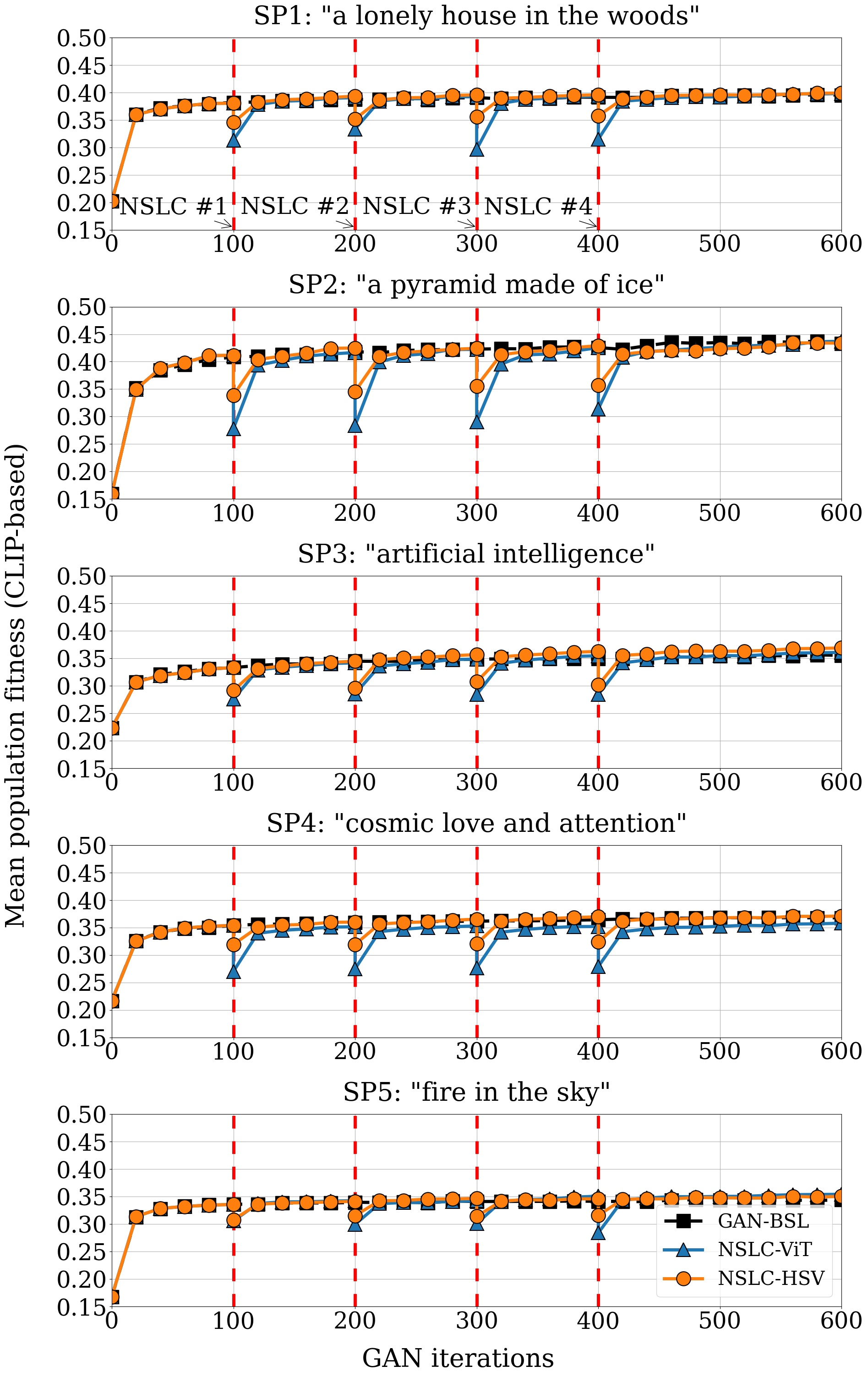} \label{fig:graph_quality}}
\hfill
\subfloat[Mean HSV diversity]
{\includegraphics[width=0.32\textwidth]{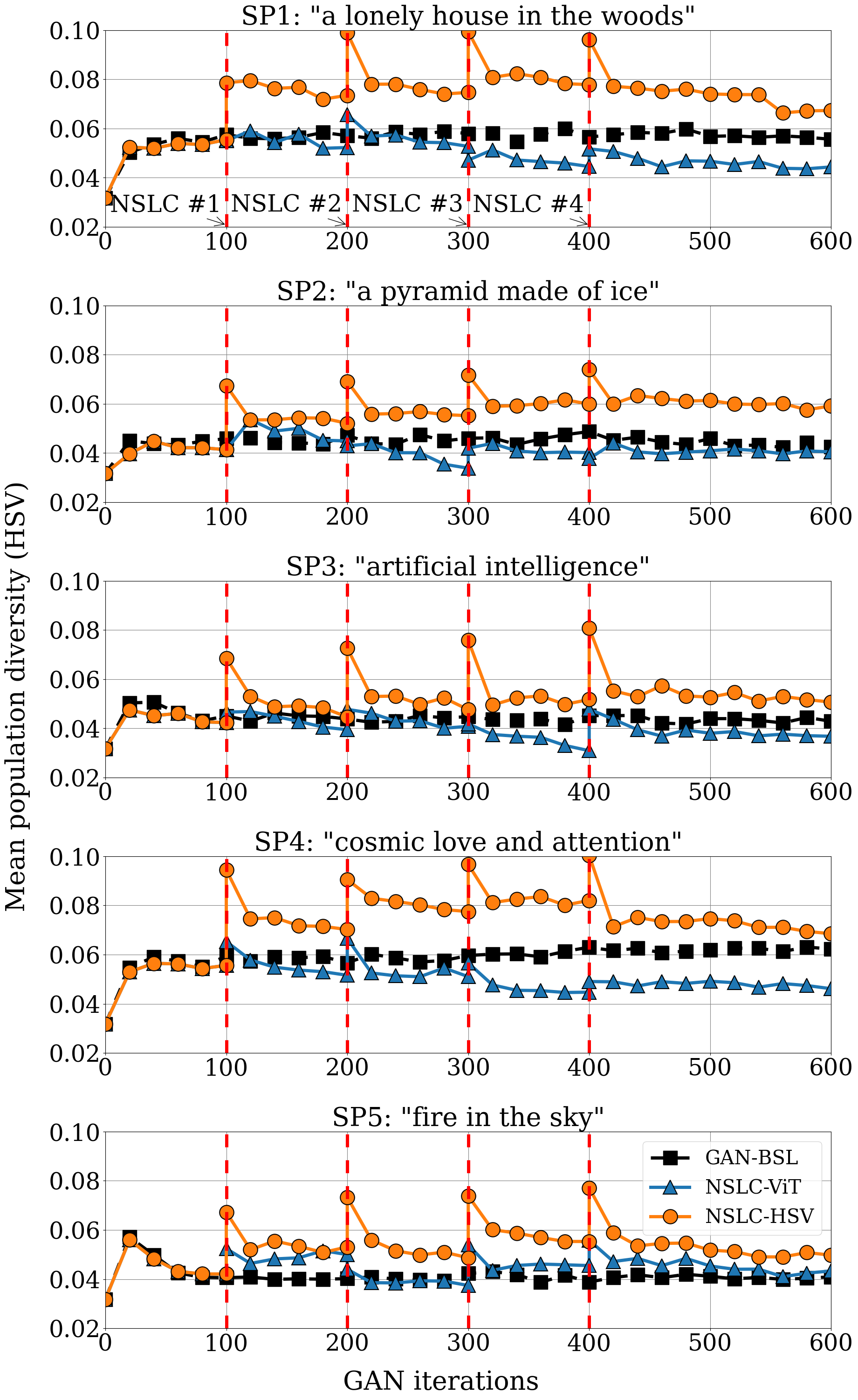} \label{fig:graph_diversity_hsv}} 
\hfill
\subfloat[Mean ViT diversity]
{\includegraphics[width=0.32\textwidth]{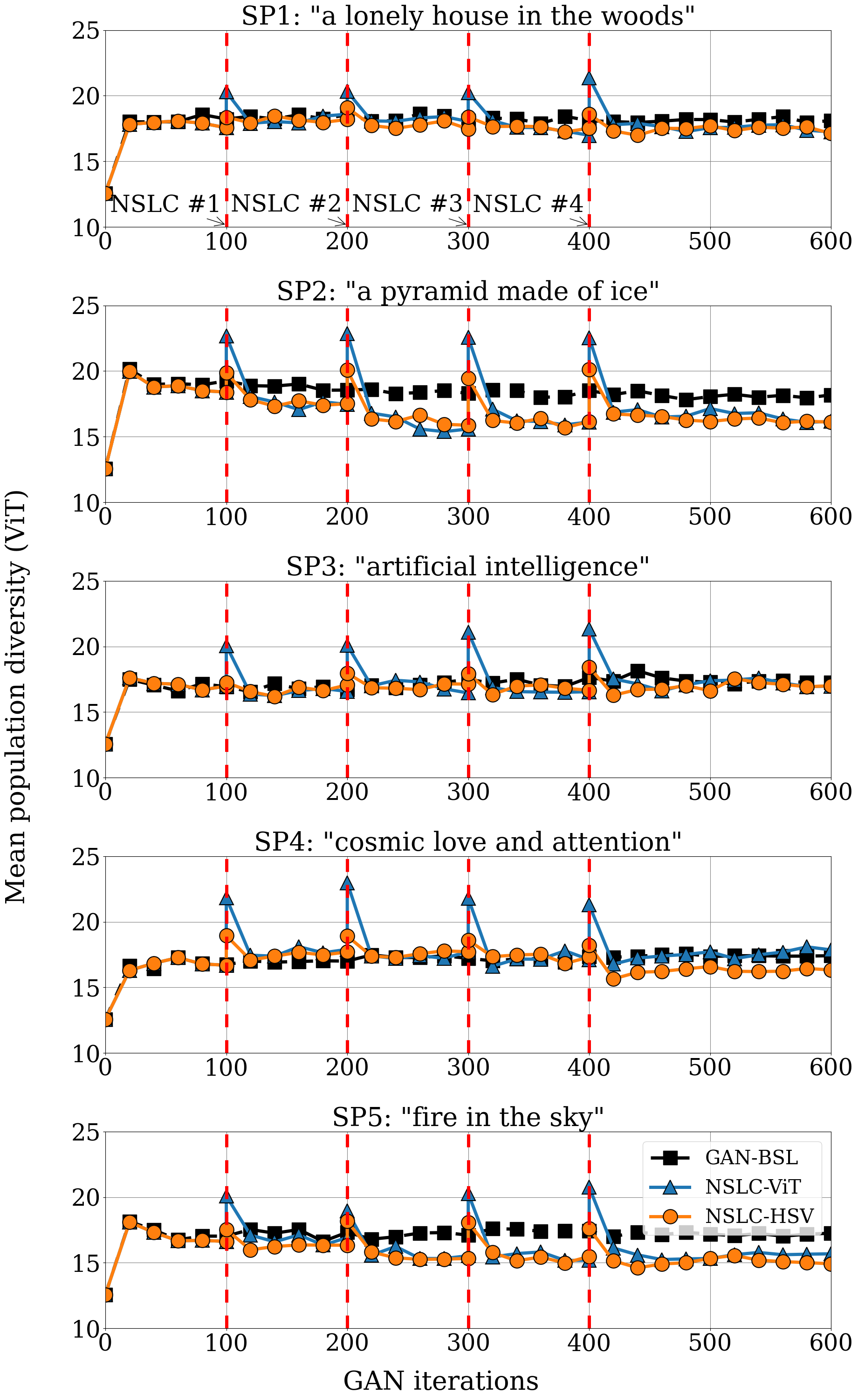} \label{fig:graph_diversity_vit}} 
\caption{Progression of the performance metrics over GAN iterations. The iterations at which evolutionary NSLC cycles were performed are marked in red.}
\label{fig:graphs}
\end{figure*}

We are equally interested in the \emph{process} followed by the algorithms tested as we are in the \emph{product} at the end of 600 iterations \cite{jordanous2016perspectives}. Therefore, Figure~\ref{fig:graph_quality} shows how the mean fitness (CLIP score) fluctuates at different GAN iterations. Evidently, with the uninterrupted GAN-BSL the algorithm increases its accuracy quickly in the first 20 iterations but then continues to slowly improve. When the process is interrupted by NSLC cycles, the evolved population's fitness drops by 12\% on average for NSLC-HSV and by  21\% for NSLC-ViT. Surprisingly, the drop is nearly as substantial when NSLC is applied at later iterations, even if the (seed) images are well-formed at that point. It is evident that after each NSLC cycle, the GAN has a similar behaviour as when facing random initial seeds and can quickly restore the CLIP score to a similar level as the GAN-BSL at the same iteration (before quickly dropping again at the next NSLC cycle). At the end of the 600 iterations, all three algorithms seem to be reaching a very similar mean fitness score, although in almost all cases both NSLC variants reach slightly higher scores than the GAN baseline (with the exception of SP4 where the mean fitness of NSLC-ViT is 2.9\% lower than GAN-BSL). Overall, NSLC-HSV seems more stable in performance, reaching on average 1.5\% higher mean fitness than GAN-BSL. By comparison, NSLC-ViT has more fluctuations between prompts and reaches an average increase of 0.7\% from the GAN-BSL mean fitness. The biggest increase in CLIP score is for SP3, where NSLC-HSV outperforms GAN-BSL by 3.9\% in terms of mean fitness.

Figures \ref{fig:graph_diversity_hsv} and \ref{fig:graph_diversity_vit} show how the mean diversity of the population fluctuates at different GAN iterations. Both image distance metrics are displayed, and the interim populations of all three methods (GAN-BSL, NSLC-ViT, NSLC-HSL) are parsed to derive these diversity values---even if they were not evolving towards that specific novelty measure. It is fairly surprising that for both image distance metrics the diversity increases during the first 20 GAN iterations. One would expect that the swift increase of the CLIP score (see Fig.~\ref{fig:graph_quality}) during those early stages would come at the cost of diversity as the images are pushed towards a generic style imposed by the manifold. For both image distance metrics, the diversity for the GAN-BSL stays fairly stable after these first few iterations, or tends to drop. This is most pronounced in SP5 for both ViT diversity and HSL diversity; we hypothesise that the (literal) prompt itself pushes images that are fairly similar in colour (red and blue) and in terms of image classification. 
Regarding the NSLC variants, we observe the reverse behaviour compared to the mean fitness plots of Fig.~\ref{fig:graph_quality}: diversity increases after each exploration cycle, at least for the distance metric targeted by novelty search. Interestingly, NSLC-HSV manages to increase both HSV diversity and ViT diversity, even if it evolves towards the former. On average, in each exploration cycle NSLC-ViT increases ViT diversity by 25\% while NSLC-HSL increases ViT diversity by 10\% (per prompt). NSLC-ViT however underperforms in terms of HSV diversity, with minor or no increases after each cycle. On the other hand, with NSLC-HSV we observe an average increase of 43\% in HSV diversity after each cycle (per prompt). 
Since images produced by NSLC are more diverse but less fit, once GAN iterations re-start the diversity quickly drops as CLIP score increases. GAN iterations after NSLC tend to lead the population to a lower ViT diversity than the GAN baseline. This behaviour is surprising, especially considering that both NSLC variants manage to increase ViT diversity during the evolutionary cycles. Even more surprising is the fact that the GAN increases ViT diversity quite dramatically when dealing with random images (at 0 iterations), but this does not seem to be the case when NSLC produces noisy images at iterations 100, 200, 300, 400. After 600 iterations, the final images of NSLC-HSV have an average of 6.3\% increase in HSV diversity compared to the GAN baseline but an average 11.5\% decrease in ViT diversity, per prompt. The final images for NSLC-ViT however are less diverse for both ViT and HSV compared to the GAN baseline (by 5.8\% and 13.7\% respectively).

\subsection{Indicative results}\label{sec:results_images}

\begin{figure}[!t]
\includegraphics[width=\columnwidth]{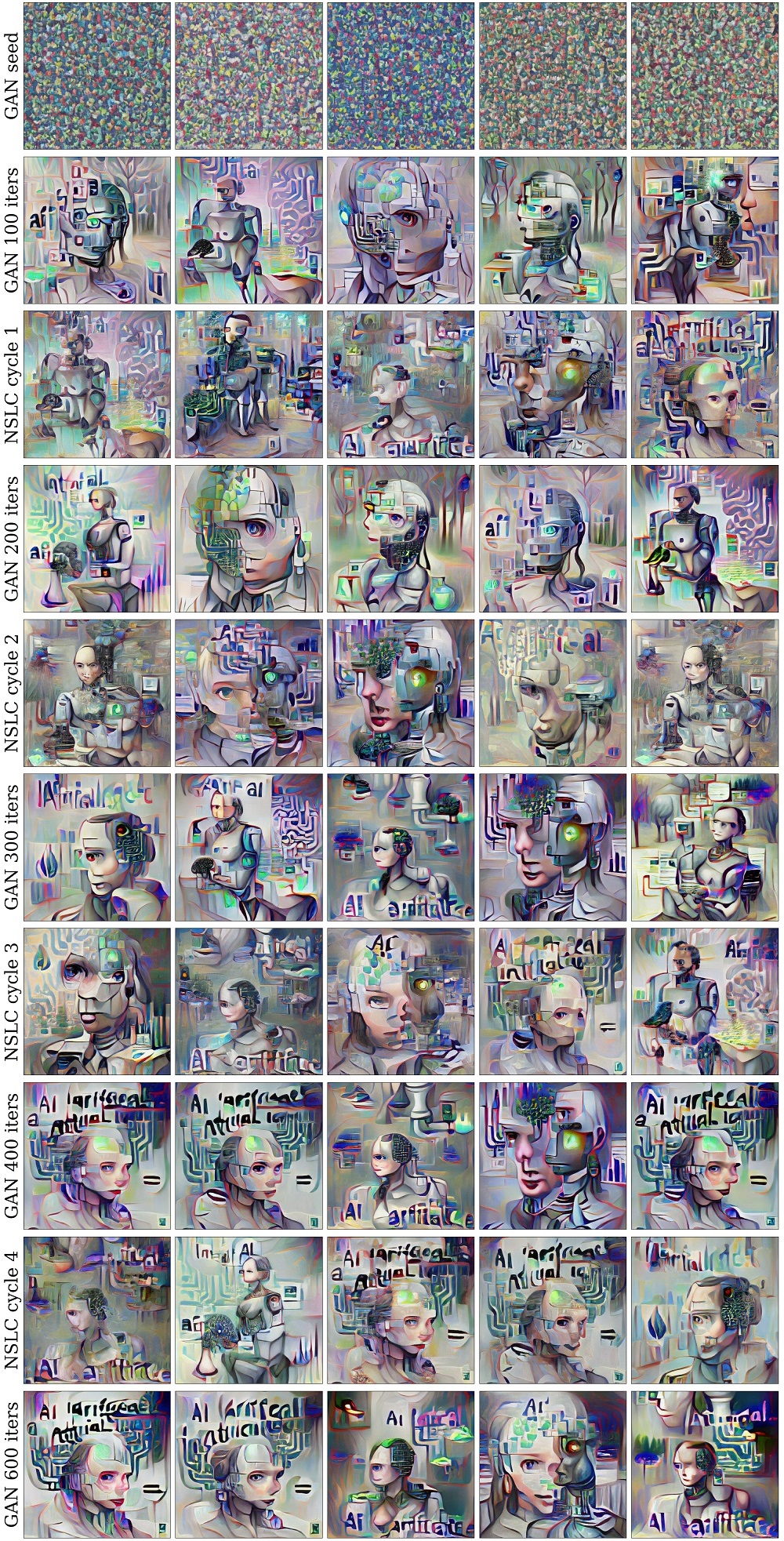}
\caption{The progression of 5 individuals (chosen for their highest nearest-neighbour HSV diversity) per population at the end of each stage in the NSLC-HSV experiment for SP3.}
\label{fig:progression}
\end{figure}

In order to better understand the process introduced in this paper, we show the most diverse images at different stages of the process. We use the HSV diversity and measure only the nearest-neighbour distance to choose the most diverse individuals at that point. Since NSLC-HSV led to more diverse individuals while maintaining comparable quality to the GAN baseline, we show results of NSLC-HSV in Figure~\ref{fig:progression}. For the purposes of brevity, we focus on SP3 since its final products have the highest increase in terms of CLIP score (3.9\% above the GAN baseline) and a good increase in HSV diversity (7\% above the GAN baseline).

It is evident that even after 100 GAN iterations, images are recognisable although their details are rough. At 100 GAN iterations, images are fairly diverse, while after the first exploration cycle new patterns are introduced (e.g. a human nose) but some images become more indistinguishable. These rough images are refined during the next GAN cycle, which results in similar-looking but crisper images. Similar rounds of exploration and refining add more details. Later NSLC cycles result in more recognisable, less noisy images. Notably, after 400 iterations the images start becoming more similar, and overarching patterns such as the introduction of the text ``artificial'' starts appearing in most images. At that stage, the last NSLC cycle does not quite manage to break these patterns and the final products at 600 iterations show more similarities than e.g. interim images at 300 iterations. We can assume that NSLC is more meaningful in early stages, and given enough time GANs will enforce their patterns even to initially novel images. Perhaps stopping the process earlier or intervening with NSLC in earlier stages (e.g. at 20 or 50 iterations rather than at 400) may better counter this drift towards dominant patterns.

\section{Discussion}\label{sec:discussion}

Our experiments investigated how quality-diversity evolutionary search applied in interim phases of a Generative Adversarial Network process can impact the creative process and products. Results show that seeding diversity in exploration cycles through NSLC can increase the diversity temporarily, but with a lesser impact in the long run as the GAN process re-asserts patterns in the corpus. While simplistic, HSV distance was shown to be better as a measure for the novelty score that guides evolution. However, observing the most diverse images in terms of HSV distance (see Fig.~\ref{fig:progression}) the differences are not as obvious to a human. It is also worth noting that this study is the first to assess the diversity of a population of random initial seeds refined through the GAN iterative process; the final products were surprisingly more diverse than expected. As a general overview, NSLC manages to increase slightly the typicality (in terms of semantic prompts) of the final generated images; however, the small increase in diversity (and only for one visual similarity metric) compared to random seeds is perhaps underwhelming considering the computational overhead of multi-objective evolution over multiple cycles throughout the process. Despite these mixed results, the notion of diversifying products of AI Art has many interesting research directions beyond the experiments reported in this paper.

While this paper explored visual diversity under different perspectives (based on models trained on labelled data and based on simple visual metrics), there are many more ways. Other measures based on deep learning, such as the Learned Perceptual Image Patch Similarity (LPIPS) metric \cite{zhang2018perceptual} can be used both as a distance metric for novelty search or as a way to evaluate the existing products' diversity. In our preliminary experiments using LPIPS for novelty score, however, the final products were not as diverse as those of the GAN baseline (in term of LPIPS). Given that HSV distance was surprisingly efficient as a novelty metric, other metrics of visual quality in the literature such as compressibility \cite{machado2015complexity} could also be explored. It should be noted that in our preliminary experiments we also explored using the binary distance\footnote{We measure binary distance as the number of items in the two images' latent vectors that were not identical at the same position.} between latent vectors (i.e. the genotype) as a measure of novelty, but the results were underwhelming. 

Beyond the distance measures, other ways of performing changes on the image during evolution can be explored. While our preliminary experiments that used recombination between two parents' latent vectors resulted in less diverse final products, better operators for mutation and recombination could lead to more creative outcomes. A potential alternative to the current random mutation of the latent vector could be to use the intermediate representation used by the GAN, which consists of a tensor of real values, in order to provide a smoother gradient if mutation is based on Gaussian noise. The disadvantage to such an approach is an increase in computational time, since this intermediate representation is much larger than the latent vector used in our current work. Another alternative would be to apply mutations on the image itself, and then allow these to be decoded into a new latent vector (rather than the reverse, which is done in the current implementation). Changes to the image can be performed as filters applied to the entire image, similar to \cite{colton2008emotionally,heath2016creating}, as local changes in a portion of the image, or taking advantage of machine-learned models such as style transfer \cite{gatys2016style}.

Extensions of this work that go beyond applying NSLC on the images themselves could provide a more direct way to demonstrate the intentionality of the computational creator. OpenAI's CLIP already offers a human understandable \cite{colton2008tripod} goal in the form of the semantic prompt. Allowing the computational creator to adapt the semantic prompt itself (e.g. by applying latent variable evolution on the semantic prompt, rather than on the image) could lead to more visually diverse images and---more importantly---to a creative process where the computational creator could change its goal and explain towards which direction it is changing (and why, presuming some objective or distance criterion). More ambitious goals in this vein could include both image adjustments (through evolution) and a corresponding change in the best semantic prompt that matches these image adjustments. Finally, the refinement could come in the form of additions to the semantic prompt, such as maximising or minimising cosine similarity with keywords (e.g. ``photorealistic'') or with intended emotional outcomes from the audience \cite{galanos2021affectgan} that are added during exploration cycles. Further work in this direction could involve a human audience assessing diversity of the resulting images, thereby highlighting how the metrics match (or not) human perception and aesthetics.

\section{Conclusion}\label{sec:conclusion}

In this work, we highlighted how what is considered today ``AI Art'' \cite{mccormack2019autonomy} largely ignores any creative dimensions except typicality \cite{ritchie2007empirical}. We explored ways of injecting novelty both in the final products and in the process of a generative adversarial network, by interspersing cycles of artificial evolution that targets both typicality and novelty as objectives. Applying several cycles of exploration between cycles of iterative refinement, we investigated how image generation driven by state-of-the-art image-language mappings can lead to more diverse outcomes. This first experiment has shown that Novelty Search with Local Competition can lead to more visually diverse results, but also highlighted that evolution applied on the code book led to more noisy interim results which forced GAN refinements to overcompensate in terms of conformity. Many extensions to the general concept of cycles of evolutionary exploration and backpropagation-based refinement in different aspects of the AI Art process (e.g. on the image level or the prompt level) can allow for a more direct and more explainable creative process.

\section{Acknowledgements}
This project has received funding from the European Union’s Horizon 2020 programme under grant agreement No 951911.

\section{Author Contributions}
Marvin Zammit prepared and carried out the reported experiments. Marvin Zammit and Antonios Liapis jointly analysed the resulting data. Marvin Zammit and Antonios Liapis each contributed to the writing in the various sections of the paper. Antonios Liapis and Georgios N. Yannakakis advised on the research direction and the text, and oversaw the implementation, analysis, and authoring process.

\bibliographystyle{iccc}
\bibliography{seeding}
\end{document}